\documentclass[letterpaper]{article} 

\ifdefined\aaaianonymous
    \usepackage[submission]{aaai2026}  
    
\else
    \usepackage{aaai2026}              
    \nocopyright
\fi

\usepackage{times}  
\usepackage{helvet}  
\usepackage{courier}  
\usepackage[hyphens]{url}  
\usepackage{graphicx} 
\def\UrlFont{\rm}  
\usepackage{natbib}  
\usepackage{caption} 
\usepackage{amsmath}
\usepackage{amsfonts}
\usepackage{amssymb}
\usepackage{booktabs}      
\usepackage{multirow}      
\usepackage{array}         
\usepackage{amssymb}
\usepackage{pifont}

\usepackage{algorithm}
\usepackage{algorithmic}

\usepackage{newfloat}
\usepackage{listings}
\DeclareCaptionStyle{ruled}{labelfont=normalfont,labelsep=colon,strut=off} 
\floatstyle{ruled}
\newfloat{listing}{tb}{lst}{}
\floatname{listing}{Listing}

\ifdefined\aaaianonymous
    \title{Hierarchical Graph Attention Network for No-Reference Omnidirectional Image Quality Assessment}
\else
    \title{Hierarchical Graph Attention Network for No-Reference Omnidirectional Image Quality Assessment}
\fi

\author{
    Hao Yang,
    Xu Zhang, 
    Jiaqi Ma, 
    Linwei Zhu, 
    Yun Zhang, 
    Huan Zhang
}
\affiliations{

    2112403134@mail2.gdut.edu.cn, zhangx0802@whu.edu.cn, jiaqi.ma@mbzuai.ac.ae,\\lw.zhu@siat.ac.cn, zhangyun2@mail.sysu.edu.cn, huanzhang2021@gdut.edu.cn
}

\usepackage{bibentry}

\begin{document}

\maketitle

\begin{abstract}

Current Omnidirectional Image Quality Assessment (OIQA) methods struggle to evaluate locally non-uniform distortions due to inadequate modeling of spatial variations in quality and ineffective feature representation capturing both local details and global context. To address this, we propose a graph neural network-based OIQA framework that explicitly models structural relationships between viewports to enhance perception of spatial distortion non-uniformity. Our approach employs Fibonacci sphere sampling to generate viewports with well-structured topology, representing each as a graph node. Multi-stage feature extraction networks then derive high-dimensional node representation. To holistically capture spatial dependencies, we integrate a Graph Attention Network (GAT) modeling fine-grained local distortion variations among adjacent viewports, and a graph transformer capturing long-range quality interactions across distant regions. Extensive experiments on two large-scale OIQA databases with complex spatial distortions demonstrate that our method significantly outperforms existing approaches, confirming its effectiveness and strong generalization capability.
\end{abstract}

\ifdefined\aaaianonymous
\else
\begin{links}
\end{links}
\fi

\begin{figure}[t]
\centering
\includegraphics[width=\linewidth]{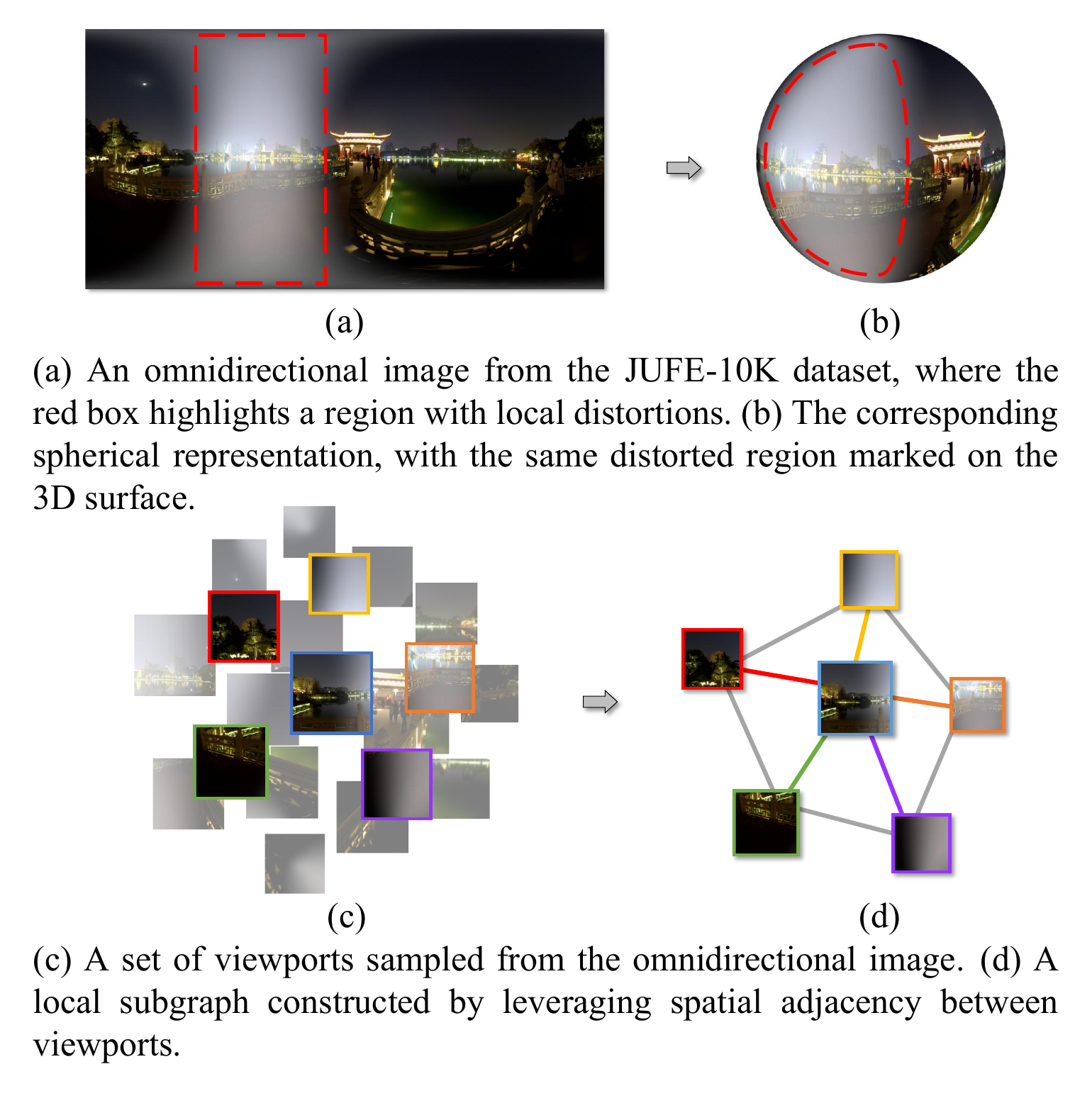}
\caption{Illustration of the visual pipeline for modeling spatially non-uniform distortions in omnidirectional images.}
\label{visual}
\end{figure}

\section{Introduction}
With the continuous advancement of imaging technology and the growing demand for immersive content, 360-degree omnidirectional images have attracted widespread attention in applications such as Virtual Reality (VR), Augmented Reality (AR), and the metaverse \cite{duan2024quick}. Compared with conventional 2D images, omnidirectional images can capture the full field of view of a scene, providing users with a more realistic and immersive visual experience. They have been widely used in domains such as online tourism, virtual exhibitions, remote conferencing, and surveillance. However, during image acquisition, stitching, projection, compression, and transmission, omnidirectional images are highly susceptible to various distortions \cite{wang2017begin}. Due to their complex spatial structure, these distortions often result in more severe quality degradation \cite{zhang2024quality}. Therefore, developing effective Omnidirectional Image Quality Assessment (OIQA) methods is of great significance for enhancing user experience and optimizing system performance \cite{xu2020state}.

Typically, OIQA methods can be categorized into two main types: Full-Reference (FR) and No-Reference (NR) approaches. FR methods rely on high-quality reference images. Early studies extended traditional 2D IQA models to the projected representations of omnidirectional images, such as applying PSNR and SSIM to Equirectangular Projection (ERP) images \cite{wang2004image}. Although these methods are relatively simple to implement, they fail to account for user perception and viewing behavior, which limits their consistency with human subjective experience. In contrast, NR methods do not require access to reference images and are more suitable for practical deployment. As a result, they have become the focus of recent research in the field.

In practical scenarios, omnidirectional images are usually stored and transmitted in the form of ERP. This projection, which maps the spherical image onto a 2D plane, introduces considerable geometric distortions, especially in polar regions, which result in noticeable inconsistencies between the projected content and human visual perception during viewing. To improve perceptual consistency, some NR methods adopt viewport-based evaluation strategies \cite{wu2023assessor360,sun2019mc360iqa,xu2020blind,zhou2021omnidirectional}, which simulate user viewing behavior by extracting local regions from different directions to estimate quality. Since viewports are closer to the actual field of human vision, these methods have demonstrated superior performance in many tasks and provide a more perception-aligned modeling strategy for OIQA.

Although viewport-based methods perform well in scenarios with uniformly distributed distortions, they face challenges when dealing with omnidirectional images that exhibit non-uniform distortion patterns \cite{fang2022perceptual}. Fine-grained perception of distortion types and severities plays a crucial role in enhancing the accuracy and robustness of image quality assessment models \cite{zhang2025perceive,zhang2025uniuir}. In such cases, the distortions are often spatially complex and unevenly distributed, leading to significant quality variations across different regions, as shown in Fig.~\ref{visual}(a)(b). These characteristics increase the difficulty of achieving consistent and accurate quality estimation.

To effectively address the unique challenges posed by spatially non-uniform distortions in omnidirectional images, we propose a no-reference quality assessment model with hierarchical graph modeling. Specifically, we first adopt a Fibonacci sphere sampling strategy \cite{saff1997distributing} to generate perceptually uniform viewports across the sphere. To extract discriminative and quality-sensitive representations, each viewport is encoded using a pre-trained Swin Transformer \cite{liu2021swin} that captures both local texture and hierarchical semantics. Recognizing the importance of spatial context, we construct a local subgraph by connecting each viewport to its $k$ nearest neighbors based on the Haversine distance, effectively preserving geometric proximity, as shown in Fig.~\ref{visual}(c)(d). A Graph Attention Network (GAT) \cite{velickovic2017graph} is employed to model localized quality dependencies among neighboring viewports. Beyond local interactions, we further incorporate a Graph Transformer to capture long-range structural correlations, allowing the model to reason across distant regions on the sphere. This unified framework enables the joint modeling of local distortion sensitivity and global structural coherence, leading to more accurate and perceptually aligned quality prediction.

The main contributions of this work are as follows:

\begin{itemize}
\item We adopt Fibonacci sphere sampling to extract a set of spatially distributed viewports from omnidirectional images, which facilitates spatially consistent information input for quality modeling.
\item We construct spherical viewport adjacency graphs based on the Haversine distance and apply a GAT-based module to model local interactions among neighboring viewports.
\item We design a Graph Transformer architecture that leverages the spatial topology of viewports to capture global quality dependencies across regions.
\item Extensive experiments conducted on two large-scale omnidirectional IQA datasets containing non-uniform distortions validate the effectiveness of our method, which outperforms existing methods in terms of both prediction accuracy and cross-dataset generalization.
\end{itemize}

\begin{figure*}[t]
\centering
\includegraphics[width=0.95\textwidth]{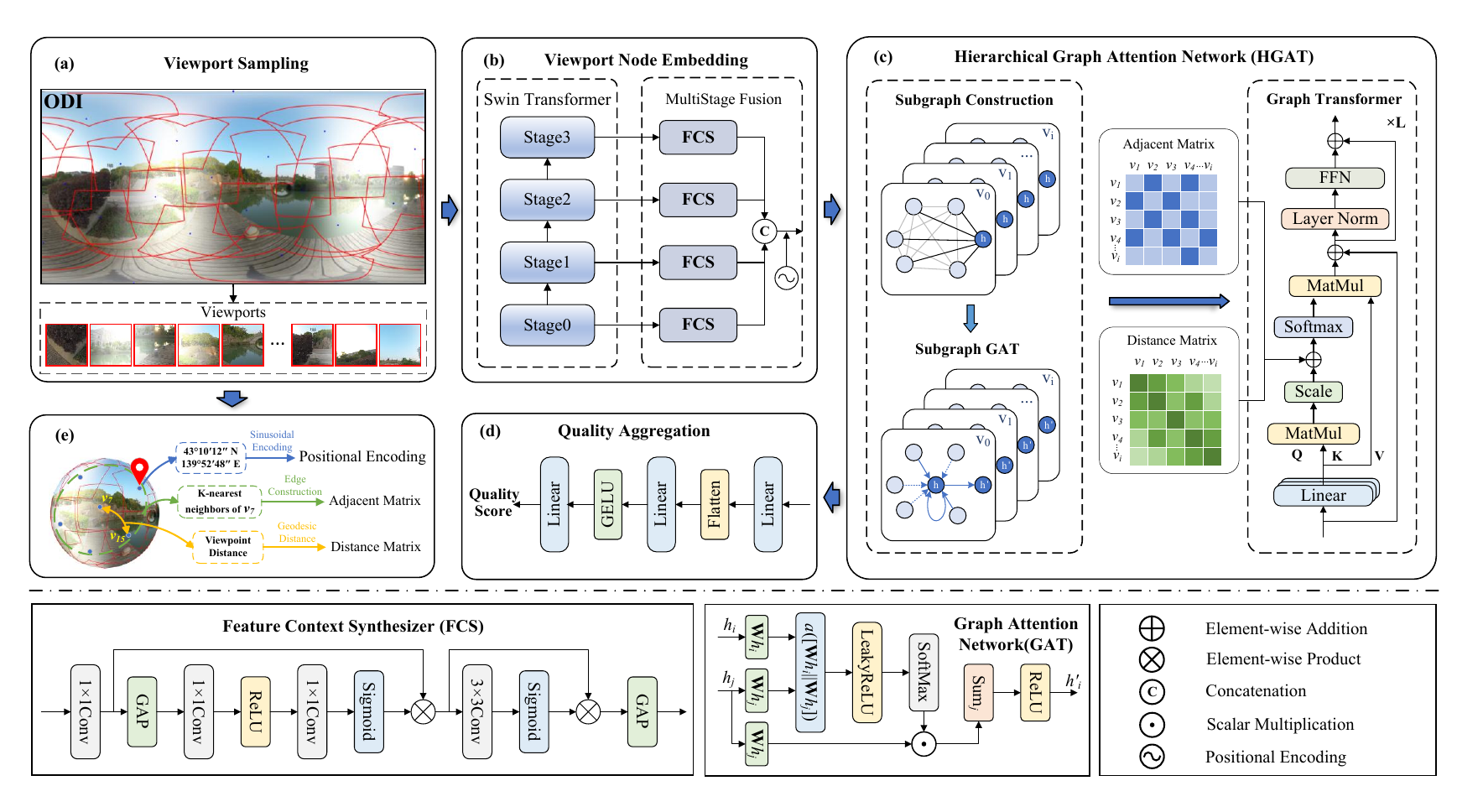}
\caption{An overview of the proposed model architecture.}
\label{fig:model}
\end{figure*}

\section{Related Work}

\subsection{OIQA models}

Full-Reference Omnidirectional Image Quality Assessment (FR-OIQA) requires the original reference image as input to measure the degree of quality degradation in the distorted image. Unlike traditional 2D images, omnidirectional images are represented on the sphere and projected into a planar format, which introduces stretching distortions in the geometric structure. As a result, traditional 2D-IQA metrics cannot be directly applied.

To address this issue, existing studies have proposed extensions to the peak signal-to-noise ratio (PSNR) metric. S-PSNR computes the error based on uniform sampling over the sphere to avoid distortion caused by planar projection \cite{yu2015framework}. WS-PSNR introduces a position-based weighting mechanism, adjusting the error according to the degree of stretching in each pixel's corresponding region to enhance robustness to spherical deformation \cite{sun2017weighted}. CPP-PSNR maps the omnidirectional image to the Craster parabolic projection space and computes quality differences in a projection domain that better aligns with human perception \cite{zakharchenko2016quality}. For extensions of the structural similarity index (SSIM), S-SSIM calculates image similarity in the spherical domain to compensate for projection error \cite{chen2018spherical}. WS-SSIM further considers spherical uniformity to optimize the weighting strategy \cite{zhou2018weighted}.
The potential of the above methods remains limited. In addition, in real-world scenarios, the reference image is often difficult to obtain, limiting the practicality of FR-OIQA.

No-reference omnidirectional image quality assessment (NR-OIQA) does not rely on the original image and predicts quality scores by analyzing features from the distorted image itself.

According to the input structure, existing NR-OIQA methods can be roughly categorized into the following three types.

Full-image-based no-reference methods directly take the ERP image as input, extract distortion features across the entire image, and predict the overall quality \cite{yang2021spatial}. However, due to the significant stretching in the polar regions of ERP images, the model is prone to being affected by structural deformation, thus affecting prediction accuracy.

Patch-based no-reference methods divide the omnidirectional image into multiple patches, extract features and predict local quality for each, and finally aggregate them into an overall score. \cite{sun2019mc360iqa} uses cube map projection (CMP) to obtain six cube faces, then uses convolutional neural networks to extract features and concatenate them for score regression. \cite{zheng2020segmented} uses segmented spherical projection (SSP) to extract bipolar and equatorial regions, and predicts image quality based on the combined features from both regions.

Viewport-based no-reference methods aim to better simulate the actual behavior of human viewing of omnidirectional images by sampling multiple viewport regions from specific directions as model input. \cite{xu2020blind} proposes a dual-branch architecture that extracts global quality features from the entire image and uses GCN to model relationships between viewports. \cite{fang2022perceptual} proposes a NR-OIQA model that incorporates users’ real viewing conditions into quality representation. \cite{wu2023assessor360} proposes a viewport-sequence-based method that predicts quality by averaging scores from several pseudo-viewport sequences.

\section{Method}

Our proposed omnidirectional image quality assessment framework is illustrated in Fig.~\ref{fig:model}. It takes a set of spatially distributed viewports as input and models their quality dependencies through a hybrid graph neural architecture to predict the overall perceptual quality. The design of each module is described in the following subsections.

\subsection{Viewport Sampling Based on Fibonacci Sphere Uniform Distribution}
\label{sec:sampling}
To obtain a spatially balanced and topologically consistent set of viewports over the omnidirectional image, we adopt a Fibonacci sphere sampling strategy \cite{saff1997distributing}. Unlike equiangular or longitude-latitude grid sampling \cite{kim2019deep}, Fibonacci sampling avoids over-concentration near the poles and yields nearly uniform point distributions on the sphere.

Given a predefined number of sampling points $n$, the $k$-th viewport location on the unit sphere is computed as follows:

\begin{equation}
\begin{aligned}
    z_k &= 1 - \frac{2k}{n - 1}, \\
    \theta_k &= \arccos(z_k), \\
    \psi_k &= 2\pi \cdot \left( \frac{k}{\varphi} - \left\lfloor \frac{k}{\varphi} \right\rfloor \right),
\end{aligned}
\end{equation}
where $\varphi = \frac{1 + \sqrt{5}}{2}$ denotes the golden ratio, and $\left\lfloor \cdot \right\rfloor$ represents the floor function, which returns the greatest integer less than or equal to the input. This formulation ensures that the azimuthal angle $\psi_k$ is constrained within the interval $[0, 2\pi)$ by discarding full rotations.

The resulting spherical coordinates $(\theta_k, \psi_k)$ are subsequently converted to geographic latitude and longitude via:

\begin{equation}
    \text{lat}_k = \theta_k \cdot \frac{180^\circ}{\pi} - 90^\circ, \quad
    \text{lon}_k = \psi_k \cdot \frac{180^\circ}{\pi} - 180^\circ.
\end{equation}

\begin{figure}[t]
\centering
\includegraphics[width=0.95\linewidth]{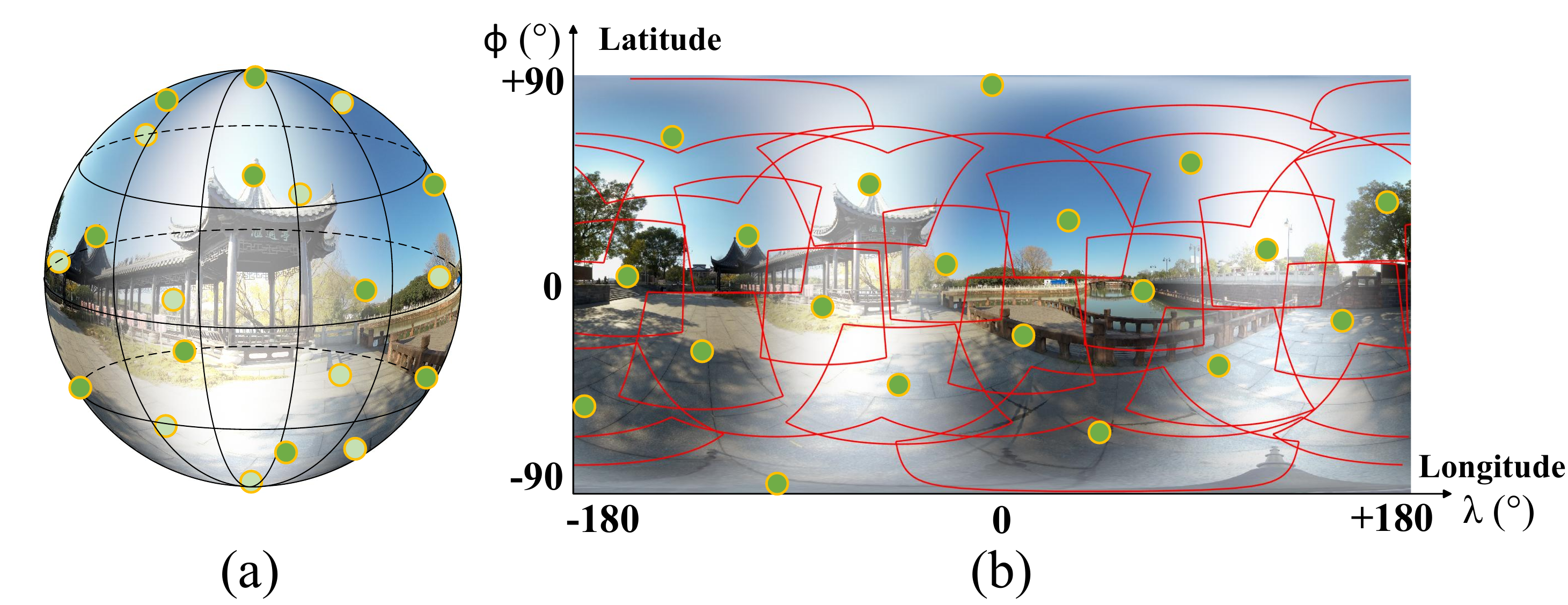}
\caption{Visualization of Fibonacci Sphere Sampling on Omnidirectional Images.}
\label{fig:fib_sampling}
\end{figure}

This strategy avoids over-concentration of points near the poles and produces a nearly uniform distribution over the sphere, making it well-suited for omnidirectional image applications. It establishes a stable and geometrically consistent spatial foundation for subsequent graph-based modeling. An illustration of the sampling pattern is shown in Fig.~\ref{fig:fib_sampling}.

\subsection{Multi-Stage Viewport Node Embedding}
\label{sec:feature_extraction}

To extract robust and quality-aware representations from each sampled viewport image, we adopt a hierarchical Swin Transformer  as the feature backbone. This architecture consists of four stages, each capturing features at different levels of abstraction with progressively reduced spatial resolution. Given an input viewport image $I \in \mathbb{R}^{3 \times H \times W}$, we obtain four feature maps $\{F_0, F_1, F_2, F_3\}$, whose spatial sizes are $\left\{ \frac{H}{8} \times \frac{W}{8}, \frac{H}{16} \times \frac{W}{16}, \frac{H}{32} \times \frac{W}{32}, \frac{H}{32} \times \frac{W}{32} \right\}$, and channel dimensions are $\{C_1, C_2, C_3, C_4\}$, respectively.

To unify and enhance these multi-scale features, we introduce a lightweight module named Feature Context Synthesizer (FCS) for each stage, consisting of Channel Attention (CA) and Spatial Attention (SA). Specifically, we first use a $1 \times 1$ convolution to project $F_i$ to a unified dimension $C$, then apply CA and SA sequentially to enhance the features. Each stage output is pooled via Global Average Pooling (GAP) to obtain a compact descriptor:
\begin{equation}
    f_i = GAP(SA(CA(F_i))) \in \mathbb{R}^{C}, \quad i = 0, 1, 2, 3.
\end{equation}

Finally, the four descriptors are concatenated to form the multi-scale viewport representation:
\begin{equation}
    h_i = Concat(f_0, f_1, f_2, f_3) \in \mathbb{R}^{4C},
\end{equation}
which is used as the node feature input for graph-based quality modeling.

This design allows the model to capture both low-level texture degradation and high-level semantic distortions in a compact and unified representation.

\subsection{Position Encoding}
\label{sec:position_encoding}

To incorporate spatial priors into the graph structure, we propose a spherical position encoding scheme for omnidirectional viewports. Each viewport’s latitude and longitude are first mapped to 3D coordinates on the unit sphere. Each coordinate axis is then independently encoded using sinusoidal functions with logarithmically spaced frequencies. For a coordinate $v$ in the 3D space, the $i$-th and $(i+1)$-th dimensions of its positional embedding are defined as:
\begin{equation}
\label{eq:sinusoid}
\begin{aligned}
PE_{(2i)}(v) &= \sin\left( \frac{v}{10000^{i / (n - 1)}} \right), \\
PE_{(2i+1)}(v) &= \cos\left( \frac{v}{10000^{i / (n - 1)}} \right),
\end{aligned}
\end{equation}
where $i \in \{0, 1, \dots, n-1\}$, and $n$ is the number of frequencies per axis. The final position encoding is obtained by concatenating the embeddings of all three axes:
\begin{equation}
p = Concat(PE(x),\, PE(y),\, PE(z)) \in R^{3 \times 2n}.
\end{equation}

This encoding captures continuous spherical geometry and is directly added to the node features, enabling the model to be aware of the spatial positions of viewports in downstream graph modules.

\subsection{Graph-Based Viewport Modeling with GAT}
\label{sec:graph_modeling}

To explicitly capture the relational dependencies among the sampled viewports, we design a three-layer GAT to model their pairwise interactions. The graph is constructed in a star-shaped fashion based on spherical geodesic distances, and each node represents a viewport with its extracted feature vector.

\paragraph{Graph Construction.} Let $\{h_1, h_2, \dots, h_V\}$ be the set of feature representations for $V$ viewports, and let $\{p_1, p_2, \dots, p_V\} \in \mathbb{R}^{V \times 2}$ denote their spherical coordinates in $(\text{lat}, \text{lon})$ form. We compute the geodesic distance between each pair of nodes using the Haversine formula:
\begin{equation}
\begin{aligned}
    d_{ij} &= 2 \cdot \arctan2\left( \sqrt{a}, \sqrt{1 - a} \right), \\
    a &= \sin^2\left( \frac{\Delta \phi}{2} \right) + \cos(\phi_i) \cos(\phi_j) \sin^2\left( \frac{\Delta \lambda}{2} \right),
\end{aligned}
\end{equation}
where $\phi_i, \phi_j$ are latitudes (in radians), $\lambda_i, \lambda_j$ are longitudes, and $\Delta \phi = \phi_j - \phi_i$, $\Delta \lambda = \lambda_j - \lambda_i$.

For each node $h_i$, we identify its $k$ nearest neighbors based on $d_{ij}$ and construct directed edges from node $i$ to its neighbors and vice versa, forming a local star-shaped graph centered at each node. Self-loops are included to preserve the node identity. The resulting graph is denoted by $G = (V, E)$ with edge index $\mathcal{E} \subseteq V \times V$.

\paragraph{GAT-based Feature Aggregation.} The constructed graph is encoded using a three-layer GAT architecture. At each layer $l \in \{1,2,3\}$, node features $X^{(l-1)} \in \mathbb{R}^{V \times C}$ are updated as:
\begin{equation}
\begin{aligned}
    H^{(l)} &= GATConv^{(l)}(X^{(l-1)}, \mathcal{E}), \\
    X^{(l)} &= \delta( LN(H^{(l)}) + X^{(l-1)} ),
\end{aligned}
\end{equation}

where $\delta(\cdot)$ denotes the ReLU activation, and $LN(\cdot)$ is layer normalization. Each $GATConv$ layer performs attention over a node’s local neighborhood. Specifically, the output for node $i$ is computed as:

\begin{equation}
\begin{aligned}
    \alpha_{ij} = 
    \frac{
        \exp\left( LeakyReLU\left( a^{\top} [W h_i \, \Vert \, W h_j] \right) \right)
    }{
        \sum_{j' \in \mathcal{N}(i)} \exp\left( LeakyReLU\left( a^{\top} [W h_i \, \Vert \, W h_{j'}] \right) \right)
    },
\end{aligned}
\end{equation}

\begin{equation}
    h_i' = \sum_{j \in \mathcal{N}(i)} \alpha_{ij} \cdot W h_j
\end{equation}

Here, $h_i$ denotes the input feature of node $i$, $W$ is a shared linear projection, $a$ is a learnable attention vector, and $\Vert$ denotes vector concatenation. $\mathcal{N}(i)$ represents the set of neighboring nodes of node $i$.

Each output feature $h_i' \in \mathbb{R}^{C}$ represents the multi-hop graph-enhanced embedding of viewport node $i$. $X \in \mathbb{R}^{V \times C}$ represents the multi-hop graph-enhanced embeddings for all viewport nodes. These embeddings are subsequently fed into a global graph transformer module, which further models long-range dependencies across distant viewport nodes for final quality prediction.

\subsection{Graph Transformer}
\label{sec:graphormer}

To capture long-range dependencies across viewports and reason about their global perceptual quality, we introduce a transformer-style architecture. Unlike classical GATs which operate on local neighborhoods, our model enables fully connected message passing over the entire graph using spatial and semantic priors as attention biases.

\paragraph{Bias Construction.}
Let $X \in \mathbb{R}^{V \times C}$ be the set of node features, where $h_i$ denotes the feature of node $i$. Two attention bias matrices are defined:

\begin{itemize}
  \item \textbf{Distance bias} $B_{\text{dist}} \in \mathbb{R}^{V \times V}$ is computed by normalizing the Haversine distance matrix $D \in \mathbb{R}^{V \times V}$, where $D$ is obtained by computing pairwise geodesic distances between nodes using the Haversine formula:
  \begin{equation}
    B_{\text{dist}}(i,j) = 1 - \frac{D_{ij} - \min(D)}{\max(D) - \min(D) + \varepsilon}
  \end{equation}
  where $D_{ij}$ is the geodesic distance between nodes $i$ and $j$, and $\varepsilon$ is a small constant for numerical stability.

  \item \textbf{Adjacency bias} $B_{\text{adj}} \in \mathbb{R}^{V \times V}$ is computed based on cosine similarity between normalized node features:
  \begin{equation}
  B_{\text{adj}}(i,j) = 
  \begin{cases}
      \frac{1}{2} \left( \frac{h_i^\top h_j}{\|h_i\|_2 \|h_j\|_2} + 1 \right), & \text{if } j \in \mathcal{N}_k(i) \\
      0, & \text{otherwise}
  \end{cases}
  \end{equation}
  where $\mathcal{N}_k(i)$ denotes the $k$ geographically nearest neighbors of node $i$.
\end{itemize}

\paragraph{Graphormer Attention.}
At each encoder layer, the input node features $X \in \mathbb{R}^{V \times C}$ are processed using a attention mechanism. We first compute the query, key, and value matrices:
\begin{equation}
    Q = X W_q,\quad K = X W_k,\quad V = X W_v,
\end{equation}
where $W_q$, $W_k$, and $W_v$ are learnable projection matrices. The attention scores are calculated with incorporated biases:
\begin{equation}
\label{eq:attn_simple}
Z = \text{softmax} \left( \frac{Q K^\top}{\sqrt{d}} + B_{\text{dist}} + B_{\text{adj}} \right) V.
\end{equation}
where $d$ is the scaling factor used in dot-product attention, and $B_{\text{dist}}, B_{\text{adj}} \in \mathbb{R}^{V \times V}$ represent distance and adjacency biases, respectively.

\paragraph{Encoder Block.}
Each Graphormer encoder layer consists of two sub-layers with residual connections:
\begin{equation}
\begin{aligned}
    X' &= X + Attention(LN_1(X)), \\
    X_{out} &= X' + FFN(LN_2(X')).
\end{aligned}
\end{equation}

where FFN is a two-layer feedforward network with GELU activation and dimensionality expansion.After stacking $L$ such encoder blocks, we obtain globally context-enhanced node features:
\begin{equation}
    X_{global} = Graphormer^{(L)}(X, B_{dist}, B_{adj}) \in R^{V \times C}.
\end{equation}

These serve as the input to the regression module for final quality prediction.

\section{Experiments}

\begin{table*}[t]
\centering
\small
\setlength{\tabcolsep}{1.5pt}
\renewcommand{\arraystretch}{0.90}

\begin{tabular}{ll|ccc|ccc|ccc|ccc|ccc}
\toprule
\multicolumn{2}{c}{} 
& \multicolumn{3}{|c|}{BD} 
& \multicolumn{3}{c|}{GB} 
& \multicolumn{3}{c|}{GN} 
& \multicolumn{3}{c|}{ST} 
& \multicolumn{3}{c}{Overall} \\
\cmidrule(lr){3-5} \cmidrule(lr){6-8} \cmidrule(lr){9-11} \cmidrule(lr){12-14} \cmidrule(lr){15-17}
Type & Models
& PLCC & SRCC & RMSE 
& PLCC & SRCC & RMSE 
& PLCC & SRCC & RMSE 
& PLCC & SRCC & RMSE 
& PLCC & SRCC & RMSE \\
\midrule

\multirow{4}{*}{FR-OIQA}
& S-PSNR     & 0.692 & 0.694 & 0.446 & 0.413 & 0.276 & 0.549 & 0.448 & 0.275 & 0.465 & 0.152 & 0.130 & 0.553 & 0.355 & 0.285 & 0.568 \\
& WS-PSNR    & 0.686 & 0.688 & 0.450 & 0.412 & 0.275 & 0.549 & 0.449 & 0.279 & 0.465 & 0.141 & 0.114 & 0.554 & 0.353 & 0.284 & 0.569 \\
& CPP-PSNR   & 0.686 & 0.688 & 0.450 & 0.412 & 0.275 & 0.549 & 0.448 & 0.274 & 0.465 & 0.140 & 0.114 & 0.554 & 0.355 & 0.285 & 0.568 \\
& WS-SSIM    & 0.415 & 0.310 & 0.562 & 0.462 & 0.312 & 0.534 & 0.448 & 0.289 & 0.465 & 0.110 & 0.055 & 0.556 & 0.388 & 0.249 & 0.560 \\
\midrule

\multirow{5}{*}{NR-IQA}
& NIQE       & 0.226 & 0.199 & 0.602 & 0.147 & 0.147 & 0.596 & 0.113 & 0.047 & 0.517 & 0.139 & 0.090 & 0.554 & 0.094 & 0.047 & 0.605 \\
& HyperIQA   & 0.267 & 0.265 & 0.595 & 0.361 & 0.353 & 0.561 & 0.256 & 0.252 & 0.502 & 0.122 & 0.102 & 0.555 & 0.201 & 0.198 & 0.595 \\
& UNIQUE     & 0.209 & 0.204 & 0.604 & 0.285 & 0.277 & 0.577 & 0.380 & 0.374 & 0.481 & 0.123 & 0.109 & 0.555 & 0.174 & 0.169 & 0.599 \\
& MANIQA     & 0.009 & 0.008 & 0.618 & 0.126 & 0.109 & 0.598 & 0.287 & 0.279 & 0.480 & 0.124 & 0.073 & 0.555 & 0.101 & 0.090 & 0.605 \\
& VCRNet     & 0.222 & 0.214 & 0.602 & 0.339 & 0.340 & 0.566 & 0.092 & 0.094 & 0.517 & 0.052 & 0.042 & 0.558 & 0.162 & 0.150 & 0.599 \\
\midrule

\multirow{6}{*}{NR-OIQA}
& MC360IQA   & 0.735 & 0.735 & 0.430 & 0.721 & 0.725 & 0.412 & 0.539 & 0.528 & 0.450 & 0.440 & 0.432 & 0.481 & 0.620 & 0.611 & 0.474 \\
& VGCN       & 0.782 & 0.770 & 0.396 & 0.443 & 0.409 & 0.533 & 0.102 & 0.087 & 0.531 & 0.207 & 0.190 & 0.524 & 0.471 & 0.377 & 0.533 \\
& Fang22     & 0.759 & 0.753 & 0.413 & 0.725 & 0.729 & 0.409 & 0.535 & 0.539 & 0.451 & 0.390 & 0.386 & 0.494 & 0.633 & 0.616 & 0.468 \\
& Assessor360& 0.727 & 0.726 & 0.435 & 0.688 & 0.696 & 0.431 & 0.733 & 0.747 & 0.363 & 0.518 & 0.516 & 0.458 & 0.694 & 0.690 & 0.435 \\
& MTAOIQA    
& 0.875 & 0.875 & 0.307 
& 0.829 & 0.831 & 0.333 
& 0.786 & 0.794 & 0.330 
& 0.689 & 0.684 & 0.388 
& 0.822 & 0.821 & 0.344 \\

& \textbf{Ours} & \textbf{0.877} & \textbf{0.877} & \textbf{0.305} & \textbf{0.842} & \textbf{0.843} & \textbf{0.321} & \textbf{0.814} & \textbf{0.821} & \textbf{0.300} & \textbf{0.744} & \textbf{0.736} & \textbf{0.358} & \textbf{0.840} & \textbf{0.840} & \textbf{0.328} \\
\bottomrule
\end{tabular}

\caption{Performance comparison of representative NR-IQA and OIQA methods on the JUFE-10K dataset. 
Here, BD, GB, GN, and ST denote brightness discontinuity, Gaussian blur, Gaussian noise, and stitching distortion, respectively.
}

\label{tab:performance}
\end{table*}

\begin{table*}[t]
\centering
\small
\renewcommand{\arraystretch}{0.90}
\setlength{\tabcolsep}{1.5pt}

\begin{tabular}{ll|ccc|ccc|ccc|ccc|ccc}
\toprule
\multicolumn{2}{c}{} 
& \multicolumn{3}{|c}{N} 
& \multicolumn{3}{|c}{L1} 
& \multicolumn{3}{|c}{L2} 
& \multicolumn{3}{|c}{G} 
& \multicolumn{3}{|c}{Overall} \\
\cmidrule(lr){3-5} \cmidrule(lr){6-8} \cmidrule(lr){9-11} \cmidrule(lr){12-14} \cmidrule(lr){15-17}
Type & Models
& PLCC & SRCC & RMSE 
& PLCC & SRCC & RMSE 
& PLCC & SRCC & RMSE 
& PLCC & SRCC & RMSE 
& PLCC & SRCC & RMSE \\
\midrule

\multirow{4}{*}{FR-OIQA}
& S-PSNR     & --    & --    & --    & 0.237 & 0.216 & 0.316 & 0.359 & 0.275 & 0.341 & --    & --    & --    & 0.302 & 0.252 & 0.343 \\
& WS-PSNR   & --    & --    & --    & 0.220 & 0.188 & 0.317 & 0.355 & 0.271 & 0.341 & --    & --    & --    & 0.295 & 0.248 & 0.344 \\
& CPP-PSNR & --    & --    & --    & 0.220 & 0.188 & 0.317 & 0.355 & 0.271 & 0.341 & --    & --    & --    & 0.295 & 0.248 & 0.344 \\
& WS-SSIM   & --    & --    & --    & 0.076 & 0.069 & 0.324 & 0.259 & 0.094 & 0.353 & --    & --    & --    & 0.223 & 0.062 & 0.351 \\
\midrule

\multirow{5}{*}{NR-IQA}
& NIQE    & 0.178 & 0.123 & 0.430 & 0.075 & 0.080 & 0.324 & 0.098 & 0.067 & 0.363 & 0.400 & 0.417 & 0.421 & 0.368 & 0.271 & 0.436 \\
& HyperIQA & 0.265 & 0.251 & 0.421 & 0.164 & 0.160 & 0.321 & 0.242 & 0.239 & 0.354 & 0.581 & 0.577 & 0.374 & 0.475 & 0.444 & 0.412 \\
& UNIQUE & 0.408 & 0.399 & 0.399 & 0.187 & 0.183 & 0.319 & 0.242 & 0.238 & 0.354 & 0.663 & 0.667 & 0.344 & 0.573 & 0.538 & 0.384 \\
& MANIQA & 0.028 & 0.034 & 0.357 & 0.004 & 0.018 & 0.363 & 0.164 & 0.155 & 0.458 & 0.296 & 0.296 & 0.416 & 0.113 & 0.098 & 0.466 \\
& VCRNet & 0.187 & 0.182 & 0.429 & 0.082 & 0.070 & 0.324 & 0.242 & 0.230 & 0.354 & 0.341 & 0.340 & 0.432 & 0.289 & 0.277 & 0.449 \\
\midrule

\multirow{6}{*}{NR-OIQA}
& MC360IQA & 0.502 & 0.462 & 0.386 & 0.446 & 0.424 & 0.273 & 0.626 & 0.625 & 0.271 & 0.782 & 0.767 & 0.281 & 0.721 & 0.710 & 0.319 \\
& VGCN         & 0.411 & 0.383 & 0.407 & 0.498 & 0.479 & 0.265 & 0.654 & 0.649 & 0.263 & 0.763 & 0.728 & 0.292 & 0.706 & 0.699 & 0.325 \\
& Fang22    & 0.553 & 0.531 & 0.372 & 0.542 & 0.522 & 0.257 & 0.673 & 0.670 & 0.257 & 0.832 & 0.798 & 0.251 & 0.769 & 0.758 & 0.293 \\
& Assessor360 & 0.678 & 0.679 & 0.328 & 0.508 & 0.479 & 0.263 & 0.663 & 0.652 & 0.260 & 0.838 & 0.811 & 0.247 & 0.790 & 0.773 & 0.281 \\
& MTAOIQA                  & 0.717 & 0.719 & 0.311 & \textbf{0.660} & \textbf{0.653} & \textbf{0.229} & 0.731 & 0.734 & 0.237 & \textbf{0.861} & \textbf{0.831} & \textbf{0.230} & 0.829 & 0.824 & 0.256 \\
& \textbf{Ours}            & \textbf{0.735} & \textbf{0.749} & \textbf{0.302} & 0.641 & 0.636 & 0.234 & \textbf{0.736} & \textbf{0.740} & \textbf{0.234} & 0.856 & 0.822 & 0.233 & \textbf{0.837} & \textbf{0.833} & \textbf{0.251} \\

\bottomrule
\end{tabular}

\caption{Detailed performance comparison of representative NR-IQA and OIQA methods on the OIQ-10K dataset. The distortion types ‘N’, ‘L1’, ‘L2’, and ‘G’ correspond to no perceptible distortion, a one distorted region, two distorted regions, and global distortion, respectively.}

\label{tab:performance_comparison}
\end{table*}

\subsection{Experiment Settings}
\label{sec:settings}

\paragraph{Datasets.} 
We conduct a comprehensive evaluation of the proposed method on two large-scale OIQA datasets, namely JUFE-10K \cite{yan2025subjective} and OIQ-10K \cite{yan2025omnidirectional}. The JUFE-10K dataset focuses on scenarios with non-uniform distortions and contains a total of 10,320 omnidirectional images with non-uniform distortions. In contrast, the OIQ-10K dataset covers a wider range of distortion types, including non-uniform distortion, global distortion, and no perceptible distortion, and consists of 10,000 omnidirectional images. To ensure fair comparisons, we follow the official train/test splits provided in the original datasets.

\paragraph{Implementation Details.}

We sample 20 viewports from each omnidirectional image using the Fibonacci sphere sampling strategy. Each extracted viewport is then resized to $224 \times 224$ pixels. To construct the graph, each viewport is connected to its $k = 5$ nearest neighbors based on geodesic distance. The feature dimension of each node is set to 768.

Both the GAT and Graph Transformer modules use 4 attention heads. The Graph Transformer encoder contains $L = 2$ layers. The model is trained using the AdamW optimizer with an initial learning rate of $1 \times 10^{-5}$, a batch size of 4, and for 30 epochs.

All experiments are conducted on a workstation equipped with an Intel\textsuperscript{\textregistered} Xeon\textsuperscript{\textregistered} Silver 4210 CPU @ 2.20\,GHz and two NVIDIA GeForce RTX 4090 GPUs.

\paragraph{Evaluation Metrics.}
To quantitatively evaluate the performance of our model, we adopt three widely used metrics in image quality assessment: Spearman Rank Correlation Coefficient (SRCC), Pearson Linear Correlation Coefficient (PLCC), and Root Mean Squared Error (RMSE). SRCC and PLCC assess the monotonic and linear relationships between predicted and ground-truth scores, respectively, while RMSE measures the absolute prediction error.

\subsection{Performance Comparison}

We compare the proposed method with a variety of representative image quality assessment models, which can be categorized into the following three types: FR-OIQA methods, including S-PSNR \cite{yu2015framework}, WS-PSNR \cite{sun2017weighted}, CPP-PSNR \cite{zakharchenko2016quality}, and WS-SSIM \cite{zhou2018weighted}; 2D NR-IQA methods, including NIQE \cite{mittal2012making}, HyperIQA \cite{su2020blindly}, UNIQUE \cite{zhang2021uncertainty}, MANIQA \cite{yang2022maniqa}, and VCRNet \cite{pan2022vcrnet}; and NR-OIQA methods, including MC360IQA \cite{sun2019mc360iqa}, VGCN \cite{xu2020blind}, Fang22 \cite{fang2022perceptual}, Assessor360 \cite{wu2023assessor360}, and MTAOIQA \cite{yan2024multitask}. The experimental results on the JUFE-10K and OIQ-10K datasets are presented in Table~\ref{tab:performance} and Table~\ref{tab:performance_comparison}, respectively. It is worth noting that, to ensure fair and valid evaluation, all results of the compared methods are directly quoted from their original publications \cite{yan2024multitask}.

Experimental results demonstrate that traditional 2D NR-IQA methods perform generally poorly on non-uniformly distorted omnidirectional images, primarily because they neglect the spherical geometric characteristics unique to panoramic content. Although some FR-OIQA methods incorporate spherical projection modeling to mitigate projection distortions, they typically fail to consider the viewport-based local perception mechanism of the human visual system during actual observation, which limits their effectiveness. In contrast, NR-OIQA methods, which place greater emphasis on simulating realistic viewing behaviors, show noticeable performance improvements.

Building upon these comparative results, our method achieves superior performance by further modeling both local and global structural characteristics. Specifically, we treat each uniformly sampled viewport on the sphere as a node in a graph. We employ a GAT to model the local relationships among viewports and adopt a Graph Transformer to capture long-range spatial dependencies. This structure-aware framework enables the effective integration of both local and global quality correlations. Benefiting from this joint modeling strategy, our method achieves significant improvements in overall prediction accuracy.

\subsection{Cross-Dataset Generalization}

To evaluate the generalization ability of the proposed method, we conduct cross-dataset experiments between JUFE-10K and OIQ-10K. Specifically, we train the model on one dataset and test it on the other, simulating real-world scenarios with distribution shifts in distortion types and image content.

As shown in Table~\ref{tab:cross_database}, when trained on JUFE-10K and tested on OIQ-10K, our method improves PLCC and SRCC by 11.9\% and 12.6\%, respectively, compared to MTAOIQA. Conversely, when trained on OIQ-10K and tested on JUFE-10K, our model achieves PLCC and SRCC improvements of 6.3\% and 7.2\%, respectively.

These results indicate that the proposed method maintains stable performance across different training-testing combinations and consistently outperforms existing approaches in both directions. The strong robustness under varying distortion distributions is attributed to the model's explicit modeling of spatial structures among viewports, which effectively captures the spatial organization of local quality degradations and enhances its alignment with human perceptual experience.

\begin{table}[t]
\centering
\small 
\renewcommand{\arraystretch}{1}
\setlength{\tabcolsep}{2.5pt} 
\begin{tabular}{l|ccc|ccc}
\toprule
\multirow{3}{*}{Model}
  & \multicolumn{3}{c|}{Train: JUFE‑10K}
  & \multicolumn{3}{c}{Train: OIQ‑10K} \\
  & \multicolumn{3}{c|}{Test: OIQ‑10K}
  & \multicolumn{3}{c}{Test: JUFE‑10K} \\
\cmidrule(lr){2-4} \cmidrule(lr){5-7}
  & PLCC & SRCC & RMSE & PLCC & SRCC & RMSE \\
\midrule
MC360IQA       & 0.290 & 0.278 & 1.063 & 0.319 & 0.253 & 0.576 \\
VGCN           & 0.426 & 0.418 & 0.415 & 0.550 & 0.517 & 0.508 \\
Fang22         & 0.274 & 0.162 & 0.441 & 0.429 & 0.366 & 0.549 \\
Assessor360    & 0.357 & 0.367 & 0.428 & 0.624 & 0.614 & 0.475 \\
MTAOIQA        & 0.468 & 0.469 & 0.405 & 0.708 & 0.696 & 0.427 \\
\textbf{Ours}  & \textbf{0.587} & \textbf{0.595} & \textbf{0.380}
                & \textbf{0.771} & \textbf{0.768} & \textbf{0.387} \\
\bottomrule
\end{tabular}
\caption{Cross-database evaluation results. Models are trained on one dataset and tested on the other.}
\label{tab:cross_database}
\end{table}

\begin{table}[t]
\centering
\small
\renewcommand{\arraystretch}{1}
\setlength{\tabcolsep}{1.8pt}
\begin{tabular}{cccc|ccc|ccc}
\toprule
\multirow{2}{*}{B} & \multirow{2}{*}{M} & 
\multirow{2}{*}{G} & \multirow{2}{*}{T} 
  & \multicolumn{3}{c|}{JUFE-10K} 
  & \multicolumn{3}{c}{OIQ-10K} \\
\cmidrule(lr){5-7} \cmidrule(lr){8-10}
  & & & & PLCC & SRCC & RMSE & PLCC & SRCC & RMSE \\
\midrule
\checkmark &   &   &   & 0.816 & 0.815 & 0.350 & 0.824 & 0.822 & 0.259 \\
\checkmark & \checkmark &   &   & 0.824 & 0.822 & 0.343 & 0.832 & 0.828 & 0.254 \\
\checkmark &   & \checkmark &   & 0.817 & 0.816 & 0.349 & 0.826 & 0.820 & 0.259 \\
\checkmark &   &   & \checkmark & 0.826 & 0.823 & 0.340 & 0.826 & 0.821 & 0.259 \\
\checkmark & \checkmark & \checkmark &   & 0.834 & 0.832 & 0.333 & 0.833 & 0.829 & 0.254 \\
\checkmark & \checkmark &   & \checkmark & 0.831 & 0.830 & 0.336 & 0.834 & 0.831 & 0.253 \\
\checkmark &   & \checkmark & \checkmark & 0.828 & 0.827 & 0.339 & 0.831 & 0.827 & 0.255 \\
\checkmark & \checkmark & \checkmark & \checkmark 
  & \textbf{0.840} & \textbf{0.840} & \textbf{0.328} 
  & \textbf{0.837} & \textbf{0.833} & \textbf{0.251} \\
\bottomrule
\end{tabular}
\caption{
Ablation study results of each component in the model on the JUFE‑10K and OIQ‑10K datasets. 
Here, B, M, G, and T correspond to the baseline model, Multi-Stage Fusion, Subgraph GAT, and Graph Transformer, respectively.
}

\label{tab:ablation}
\end{table}

\subsection{Ablation Study}

To evaluate the effectiveness of each core component in our architecture, we conduct ablation studies on the JUFE-10K and OIQ-10K datasets. The baseline model utilizes only the final-stage features from the Swin Transformer. We then incrementally introduce the following modules: Multi-Stage Fusion, Subgraph GAT, and Graph Transformer, and assess their individual and combined contributions.

As shown in Table~\ref{tab:ablation}, adding any single module consistently improves performance across both datasets, highlighting the benefits of multi-scale feature fusion, local relational modeling, and global structure learning. Combining any two modules yields further gains, while integrating all three achieves the best overall performance. These results confirm that multi-stage fusion enhances detail sensitivity, GAT captures local quality variations among adjacent viewports, and the Graph Transformer models long-range spatial dependencies. Their synergy significantly boosts the model’s ability to perceive and predict spatially non-uniform distortions, demonstrating the importance of combining complementary mechanisms to enhance structural perception of omnidirectional content.

We further analyze the impact of different viewport sampling strategies on performance, as shown in Table~\ref{tab:viewport_performance}. Under the same full-coverage condition, our method outperforms spherical sampling~\cite{zhou2021no,xu2019quality}, suggesting that our viewpoint distribution is more beneficial for graph modeling. A more balanced and structure-aware sampling pattern provides a stronger foundation for both node feature quality and graph topology.

Equatorial sampling~\cite{yan2024multitask} performs worse on JUFE-10K but relatively better on OIQ-10K. This discrepancy stems from dataset characteristics: JUFE-10K contains only spatially non-uniform distortions, where equator-only sampling fails to cover all regions, leading to incomplete distortion-aware representation. In contrast, OIQ-10K includes both non-uniform and globally uniform distortions, as well as perceptually distortion-free images. For images with consistent quality across the sphere, equatorial viewports can sufficiently reflect global quality. These results highlight that no single sampling strategy is universally optimal, and its effectiveness depends heavily on the spatial distribution and perceptual nature of distortions.

Overall, our findings demonstrate that both architecture design and sampling strategy play critical roles in the performance of omnidirectional image quality assessment.

\begin{table}[t]
\centering
\small
\renewcommand{\arraystretch}{1}
\setlength{\tabcolsep}{1.8pt}
\begin{tabular}{l|ccc|ccc}
\toprule
\multirow{2}{*}{Sampling Strategy}
  & \multicolumn{3}{c|}{JUFE‑10K}
  & \multicolumn{3}{c}{OIQ‑10K} \\
\cmidrule(lr){2-4} \cmidrule(lr){5-7}
  & PLCC & SRCC & RMSE & PLCC & SRCC & RMSE \\
\midrule
Spherical sampling    & 0.832 & 0.832 & 0.335 & 0.833 & 0.828 & 0.254 \\
Equatorial sampling   & 0.810 & 0.809 & 0.355 & 0.834 & 0.832 & 0.253 \\
Saliency sampling   & 0.817 & 0.818 & 0.348 & 0.815 & 0.813 & 0.266 \\
\textbf{Ours}       & \textbf{0.840} & \textbf{0.840} & \textbf{0.328} 
  & \textbf{0.837} & \textbf{0.833} & \textbf{0.251} \\
\bottomrule
\end{tabular}
\caption{Performance comparison of different viewport sampling strategies.}
\label{tab:viewport_performance}
\end{table}

\section{Conclusion}
In this paper, we propose a novel FR-OIQA method to effectively model spatially non-uniform distortions. By integrating spatial positions and content features, we construct a graph structure among viewports to capture complex quality dependencies. Multi-level viewport descriptors are designed to align with the hierarchical perception of the human visual system. A GAT is used for local interaction modeling, while a Graph Transformer captures long-range structural dependencies. Experiments on two benchmark OIQA datasets demonstrate that our method achieves superior accuracy and cross-dataset generalization. In future work, we plan to incorporate user viewing behavior and extend the framework to immersive video quality assessment.

\end{document}